\documentclass[letterpaper, 10 pt, conference]{ieeeconf}

\IEEEoverridecommandlockouts                              

\let\labelindent\relax

\usepackage{times}
\usepackage{amsmath,amsfonts}
\usepackage{algorithmic}
\usepackage{algorithm}
\usepackage{array}
\usepackage{balance}
\usepackage[dvipsnames,table]{xcolor}
\usepackage{textcomp}
\usepackage{stfloats}
\usepackage{url}
\usepackage{verbatim}
\usepackage{graphicx}
\usepackage{todonotes}
\usepackage{booktabs} 
\usepackage{multirow}
\usepackage{multicol}
\usepackage{bbm} 
\usepackage{mathtools}
\usepackage{pgf} 
\usepackage{etoolbox} 
\usepackage{enumitem} 
\usepackage{amssymb}
\usepackage{underscore}

\usepackage{hyperref} 
\usepackage{lipsum} 
\usepackage[mode=match]{siunitx} 
\usepackage{mathtools}
\usepackage{xspace}
\usepackage[acronym, nonumberlist, nogroupskip]{glossaries-extra}

\usepackage{pbox}

\usepackage{threeparttable}
\usepackage{tablefootnote}

\usepackage{tikz}

\usepackage[normalem]{ulem}
\usepackage{gensymb}

\usepackage{color}
\definecolor{ColorLightCyan}{rgb}{0.88,1,1}
\definecolor{ColorLightTurquoise}{rgb}{0.5, 1, 0.8}
\definecolor{ColorOrange}{rgb}{1.0, 0.7, 0.0}
\definecolor{ColorTrajBlue}{rgb}{0,0.5,1.0}
\definecolor{ColorTrajOrange}{rgb}{0.87,0.56,0.01}

\hypersetup{
colorlinks=true
,linkcolor=NavyBlue
,citecolor=NavyBlue
,urlcolor=NavyBlue
,citebordercolor=NavyBlue
}

\def\figref#1{Fig.~\ref{#1}}
\def\tabref#1{Tab.~\ref{#1}}
\def\eqref#1{Eq.~(\ref{#1})}

\usepackage{amssymb}
\usepackage{pifont}

\usepackage{tikz}

\newcolumntype{L}[1]{>{\raggedright\let\newline\\\arraybackslash\hspace{0pt}}m{#1}}
\newcolumntype{C}[1]{>{\centering\let\newline\\\arraybackslash\hspace{0pt}}m{#1}}
\newcolumntype{R}[1]{>{\raggedleft\let\newline\\\arraybackslash\hspace{0pt}}m{#1}}

\usepackage[edges]{forest}

\definecolor{folderbg}{RGB}{124,166,198}
\definecolor{folderborder}{RGB}{110,144,169}

\newlength\Size
\tikzset{
  folder/.pic={
    \filldraw [draw=folderborder, top color=folderbg!50, bottom color=folderbg] (-1.05*\Size,0.2\Size+3.5pt) rectangle ++(.75*\Size,-0.2\Size-2pt);
    \filldraw [draw=folderborder, top color=folderbg!50, bottom color=folderbg] (-1.15*\Size,-\Size) rectangle (1.15*\Size,\Size);
  },
  file/.pic={
    \filldraw [draw=folderborder, top color=folderbg!5, bottom color=folderbg!10] (-\Size,.4*\Size+2pt) coordinate (a) |- (\Size,-1.2*\Size) coordinate (b) -- ++(0,1.6*\Size) coordinate (c) -- ++(-2pt,2pt) coordinate (d) -- cycle (d) |- (c) ;
  },
}
\forestset{
  declare autowrapped toks={pic me}{},
  declare boolean register={pic root},
  pic root=0,
  pic dir tree/.style={
    for tree={
      folder,
      font=\ttfamily,
      grow'=0,
    },
    before typesetting nodes={
      for tree={
        edge label+/.option={pic me},
      },
      if pic root={
        tikz+={
          \pic at ([xshift=\Size].west) {folder};
        },
        align={l}
      }{},
    },
  },
  pic me set/.code n args=2{
    \forestset{
      #1/.style={
        inner xsep=2\Size,
        pic me={pic {#2}},
      }
    }
  },
  pic me set={directory}{folder},
  pic me set={file}{file},
}
\usepackage{graphicx,multirow,booktabs,makecell}
\newcommand{\rotlab}[1]{\rotatebox[origin=c]{90}{\makecell[c]{#1}}}
\usepackage{mwe}

\usepackage{caption}            
\title{\LARGE \bf
ROOM: A Physics-Based Continuum Robot Simulator for Photorealistic Medical Datasets Generation}

\author{Salvatore Esposito$^{1}$, Matías Mattamala$^{1}$, Daniel Rebain$^{2}$, Francis Xiatian Zhang$^{1}$, \\
Kevin Dhaliwal$^{1}$, Mohsen Khadem$^{1}$, and Subramanian Ramamoorthy$^{1}$
}

\begin{document}

\twocolumn[{
\renewcommand\twocolumn[1][]{#1}
\maketitle
\begin{center}
\captionsetup{font=footnotesize}
\includegraphics[width=1.0\linewidth]{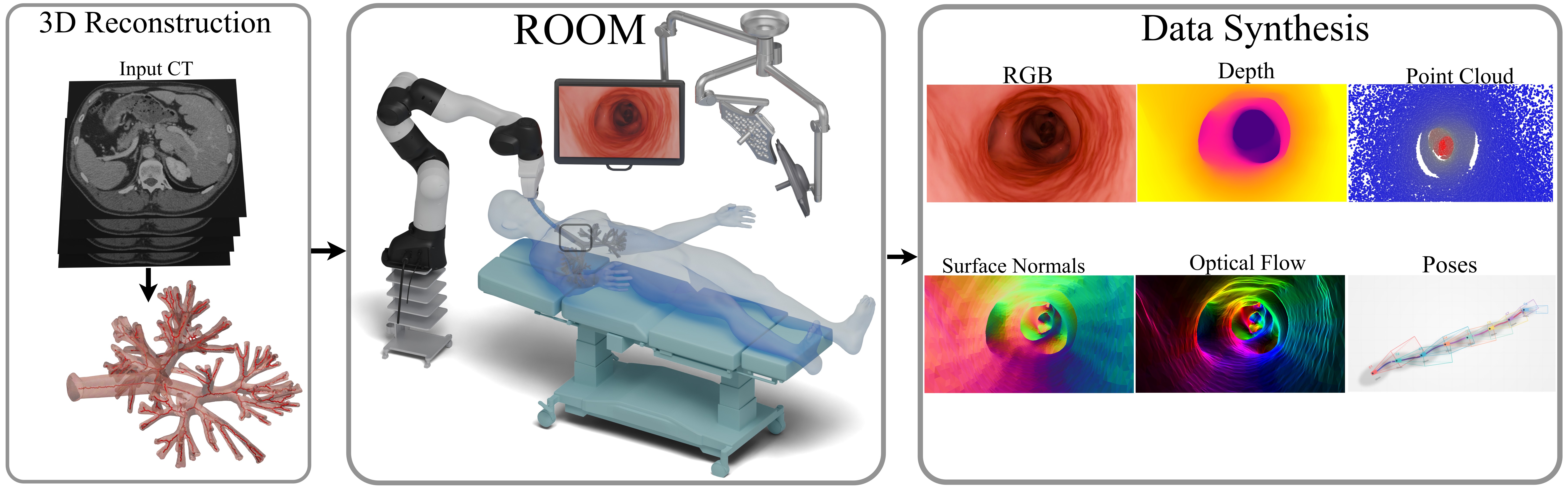} 
\captionof{figure}{{\textbf{ROOM framework overview}. Given patient CT scans (left), our pipeline reconstructs accurate 3D lung models and extracts medial axis trajectories, enabling physics-based continuum robot simulation to generate photorealistic multi-modal sensor data (right). This includes RGB images with realistic noise and lighting, metric depth maps, surface normals, optical flow, point clouds, and ground-truth poses, for different medical robotics applications.}
}
\label{fig:teaser-fig}
\end{center}
}]

\begingroup
  \renewcommand\thefootnote{}\footnote{
  \noindent$^{1}$University of Edinburgh, UK. 
  $^{2}$University of British Columbia, Canada. 
  This work was supported by a UKRI Turing AI World Leading Researcher Fellowship on AI for Person-Centred and Teachable Autonomy (grant EP/Z534833/1)
  }
  \addtocounter{footnote}{-1}
\endgroup

\thispagestyle{empty}
\pagestyle{empty}

\begin{abstract}
Continuum robots are advancing bronchoscopy procedures by accessing complex lung airways and enabling targeted interventions. However, their development is limited by the lack of realistic training and test environments: Real data is difficult to collect due to ethical constraints and patient safety concerns, and developing autonomy algorithms requires realistic imaging and physical feedback.
We present ROOM (Realistic Optical Observation in Medicine), a comprehensive simulation framework designed for generating photorealistic bronchoscopy training data. By leveraging patient CT scans, our pipeline renders multi-modal sensor data including RGB images with realistic noise and light specularities, metric depth maps, surface normals, optical flow and point clouds at medically relevant scales. 
We validate the data generated by ROOM in two canonical tasks for medical robotics: multi-view pose estimation and monocular depth estimation, demonstrating diverse challenges that state-of-the-art methods must overcome to transfer to these medical settings. Furthermore, we show that the data produced by ROOM can be used to fine-tune existing depth estimation models to overcome these challenges, also enabling other downstream applications such as navigation.
We expect that ROOM will enable large-scale data generation across diverse patient anatomies and procedural scenarios that are challenging to capture in clinical settings. Code and data: \url{https://iamsalvatore.io/room/}.
\end{abstract}

\section{INTRODUCTION}
Continuum robots have emerged as an innovative technology in minimally invasive surgery, with bronchoscopy representing one of the most promising applications. These flexible, cable-driven systems can navigate the intricate branching networks of human airways with unprecedented dexterity, enabling precise drug delivery, tissue sampling, and diagnostic imaging in lung regions previously inaccessible to rigid instruments~\cite{continuum_medical_robots}. Continuum robots can enable early intervention in peripheral lung nodules, targeted chemotherapy delivery, and real-time biopsy guidance, significantly improving patient outcomes in pulmonary medicine.

Nevertheless, the development of autonomous navigation algorithms for continuum robot bronchoscopy faces data-related limitations. Clinical data collection is inherently constrained by patient safety protocols, ethical review processes, and the high costs associated with experimental procedures. More fundamentally, the individualised nature of human anatomy means that effective algorithms must generalise across diverse airway geometries while maintaining millimetre-level precision~\cite{continuum_medical_robots}. 
Synthetic data generation has demonstrated remarkable success in addressing similar challenges across robotics applications from autonomous driving to visual SLAM~\cite{kubric_ref, tartanair_ref}. In the medical context, some recent efforts have focused on data generation for colonoscopy, as done by the SimCol3D Challenge~\cite{simcol3d}, or on simulation frameworks for surgical procedures~\cite{orbit_surgical_2024}. However, bronchoscopy procedures require anatomical fidelity, procedure-specific lighting conditions, as well as specific kinematics and sensor modalities calibrated to clinical scales. 

\begin{figure*}[ht]
  \centering
  \captionsetup{font=footnotesize}
  \includegraphics[width=0.98\linewidth, trim=0pt 0pt 0pt 0pt, clip]{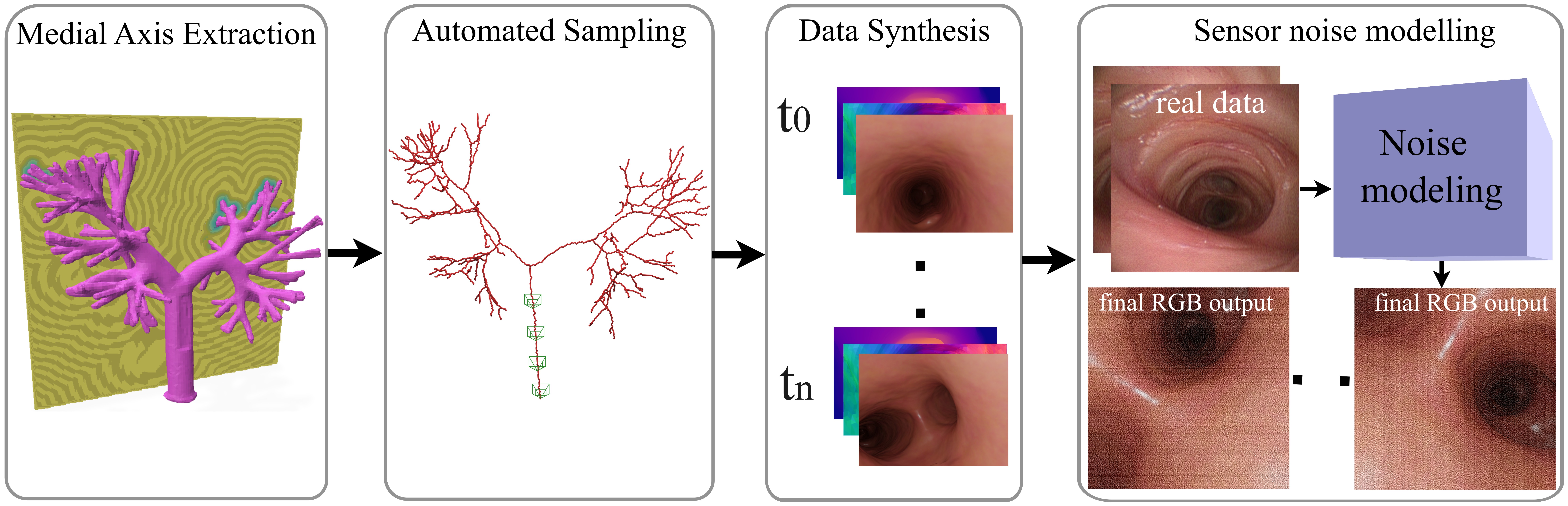}
  \caption{\textbf{ROOM data generation pipeline}. The system consists of four main stages: (1) Medial Axis Extraction from segmented CT lung models, (2) Automated Sampling along skeletal branches with higher density at bifurcations and high-curvature regions, (3) Data Synthesis generating synchronized multi-modal sensor streams from $t_0$ to $t_n$ timesteps, and (4) Sensor Noise Modeling applying realistic noise characteristics matching real bronchoscopy imagery through frequency-domain analysis.}
  \label{fig:pipeline}
  \vspace{-10pt}
\end{figure*}

In this paper, we introduce ROOM (Realistic Optical Observation in Medicine), a simulation framework engineered for continuum robot bronchoscopy applications. ROOM provides the first fully automated pipeline that transforms patient CT scan data into extensive synthetic training datasets while preserving the geometric constraints and visual characteristics essential for medical navigation tasks inside the vessels and airways of the anatomical structures. Our system generates photorealistic multi-modal sensor data, including RGB imagery with realistic noise, metric depth maps, surface normals, point clouds, and optical flow, all calibrated to the millimetre scales typical of bronchoscopy procedures, as shown in \figref{fig:pipeline}. By enabling large-scale data generation across diverse patient anatomies and challenging procedural scenarios, ROOM can facilitate the development of robot bronchoscopy without the constraints of clinical data collection.
The primary contributions of this work are:

\begin{itemize}[leftmargin=10pt]
\item ROOM, a realistic simulation framework designed for continuum robot bronchoscopy to generate synthetic data at medically-relevant scales.
\item A photorealistic rendering pipeline that considers endoscopic lighting conditions, tissue surface properties, and data-driven sensor models.
\item Validation of the synthetic data produced by ROOM in medically-relevant tasks, such as multi-view pose estimation and monocular depth estimation.
\item Demonstration of additional applications such as monocular depth fine-tuning and visual navigation.
\item Open-source release of ROOM for benefit of the community at \url{https://iamsalvatore.io/room/}.
\end{itemize}

\section{Related Work}

\noindent\textbf{Medical Robotics Simulators.}
Specialised simulation platforms for medical robotics have primarily targeted surgical training and haptic feedback~\cite{sofa_framework, orbit_surgical_2024, tmtdyn_2020}, offering real-time interaction but simplified visual rendering insufficient for training sim-to-real vision systems~\cite{surgical_sim_ref, endo_sim_ref}.
Recent neural rendering and GPU-accelerated platforms such as ORBIT-Surgical~\cite{orbit_surgical_2024} achieve fast real-time rendering for surgery and endoscopy simulation~\cite{endogaussian_2024, endora_2024}, yet are not designed for large-scale dataset generation or multi-modal sensor output (depth, optical flow, surface normals) required for navigation and depth estimation.
In colonoscopy, SimCol3D~\cite{simcol3d} introduced a Unity-based synthetic data pipeline for 3D reconstruction, pose estimation, and monocular depth estimation. However, bronchoscopy presents additional appearance and geometric degeneracies compared to the texture- and geometry-rich colon environment, necessitating advanced rendering techniques such as path tracing and BSDF shaders that ROOM integrates within its pipeline.

\noindent\textbf{Continuum Robot Bronchoscopy Systems.}
Continuum robots have shown significant potential in bronchoscopy, with clinical studies demonstrating improved diagnostic accuracy through flexible navigation of complex airway geometries~\cite{continuum_medical_robots, contact_aided_continuum}.
Prior efforts have focused on odometry and localisation: PANS~\cite{pans_2024} demonstrated 6-DOF pose tracking without external sensors via Monte-Carlo localisation given a prior lung map, while Deng et al.~\cite{deng2023dataset} introduced an ex-vivo dataset for evaluating monocular visual odometry in map-free settings. While we do not target these specific tasks, we show how ROOM-generated data supports multi-view pose estimation.
The ultimate goal of continuum bronchoscopy robots is autonomous (or semi-autonomous) navigation for localised procedures. Prior work has acquired reference trajectories in simulation~\cite{bronchopose_2022} or from real data~\cite{bm_broncholc_2024, uaal_dataset_2024}, while more recent approaches learn navigation policies via reinforcement learning in simulation~\cite{zhang_aicopilot_2024, bronchocopilot_2024}. However, these decouple physics simulation from photorealism, limiting policy performance. ROOM bridges this gap with a unified framework for visually-accurate data collection in physically-realistic settings.

\section{Method}

\subsection{Overview}

\begin{figure}[t]
  \centering
  \captionsetup{font=footnotesize}
  \includegraphics[width=1\linewidth]{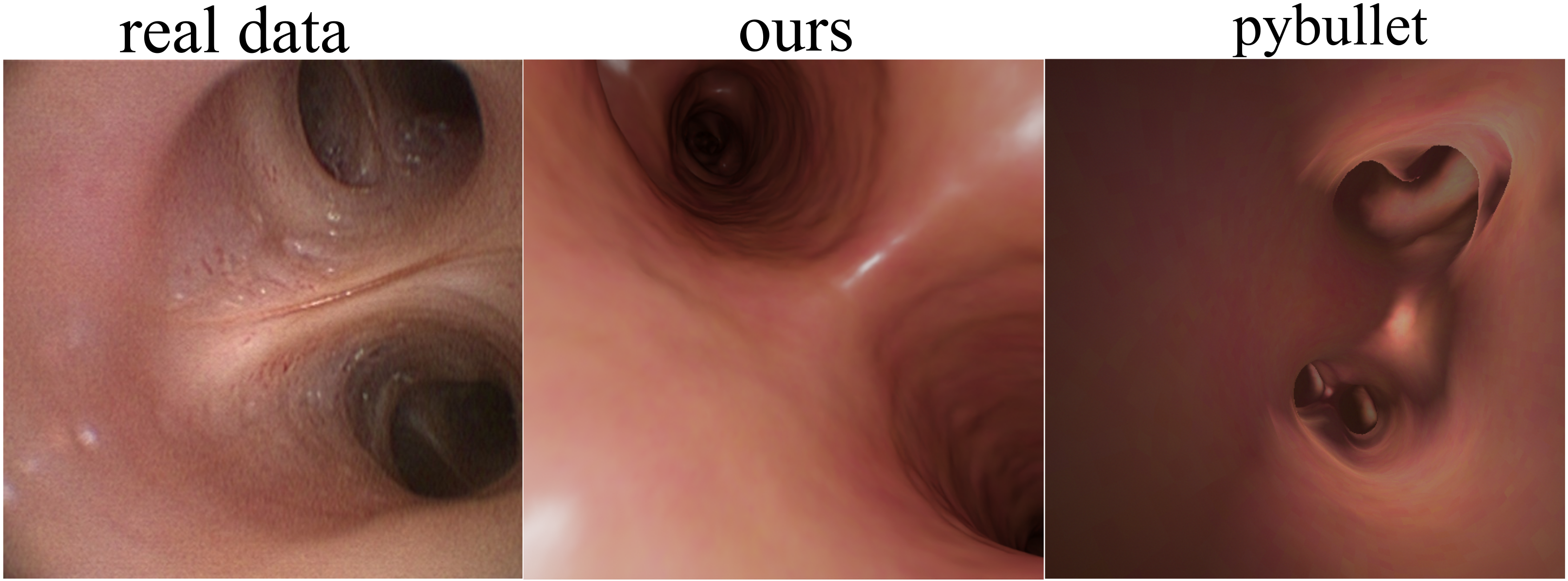}
  \caption{\textbf{Visual comparison of ROOM outputs compared to real data.} \textit{Left:} Real bronchoscopy data captured from a continuum robot showing specular highlights from wet mucosal surfaces and directional lighting. \textit{Center:} ROOM's photorealistic rendering using Blender's path tracing with Principled BSDF shaders, accurately reproducing tissue surface properties and lighting conditions. \textit{Right:} Naive PyBullet-based rendering lacking photorealistic materials and lighting.}
  \label{fig:comparison_rgb}
\end{figure}

ROOM provides a comprehensive simulation framework for generating photorealistic bronchoscopy training data using continuum robots. The system comprises four components: (1) continuum robot modelling with realistic kinematic constraints, (2) physics simulation with calibrated tissue interactions, (3) anatomical reconstruction and trajectory planning, and (4) photorealistic rendering with endoscopic artifacts. We describe these components below.

\subsection{Continuum Robot Modelling}
\label{sec:softrobot}

\begin{figure}[t]
  \centering
  \captionsetup{font=footnotesize}
  \includegraphics[width=0.95\linewidth]{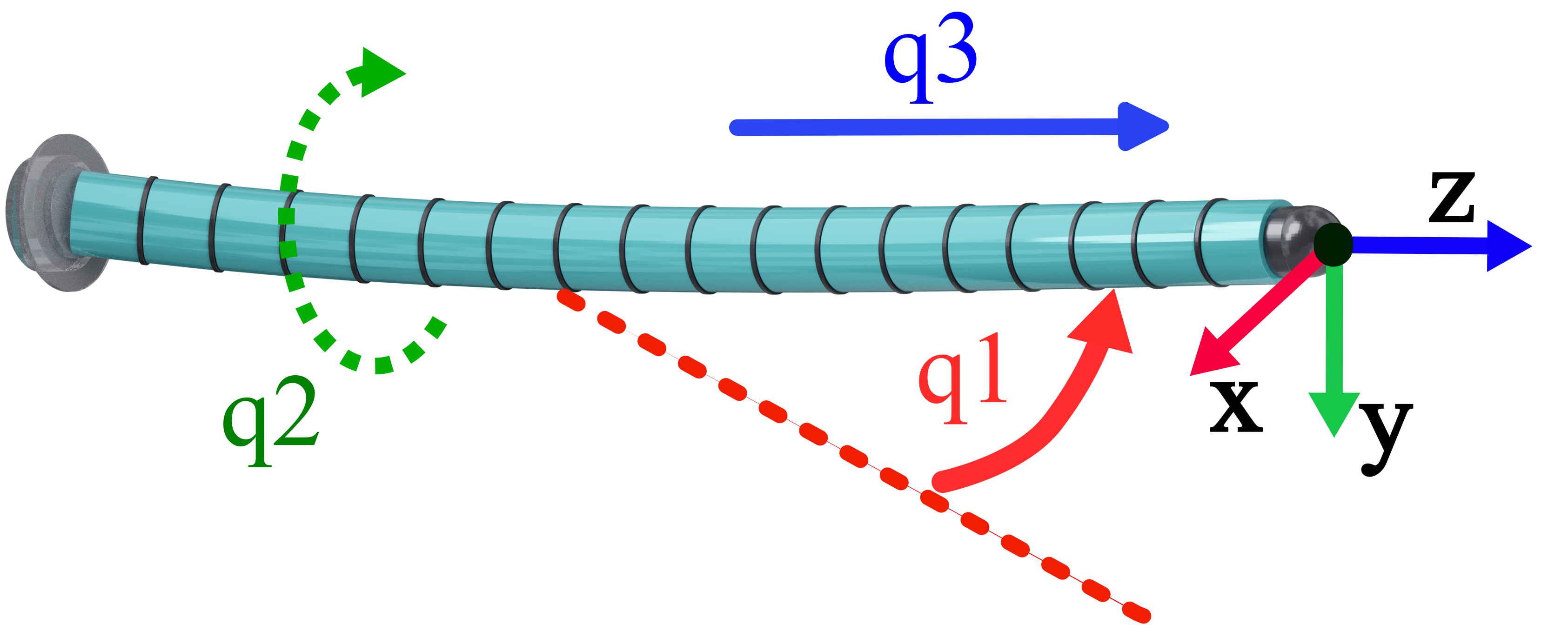}
  \caption{\textbf{Continuum robot model used in ROOM simulation}. The bronchoscope is modelled as a flexible, cable-driven continuum robot with constant curvature bending and three degrees of freedom: tendon actuation for bending curvature (q1), axial rotation for bending plane (q2), and linear insertion depth (q3). The physics-based simulation incorporates realistic friction models, actuator noise, and collision dynamics calibrated to clinical bronchoscope behaviour.}
  \label{fig:robot-model}
\end{figure}

The bronchoscope is modelled as a cable-driven continuum robot using a reduced-order piecewise-constant-strain approximation (\figref{fig:robot-model}). The model exposes three DoF aligned with clinical control interfaces: antagonistic tendon differential displacement for bending ($q_1 \in [-0.008, 0.008]~\mathrm{m}$), axial rotation selecting the bending plane ($q_2 \in \mathbb{R}~\mathrm{rad}$), and linear insertion depth ($q_3 \in \mathbb{R}~\mathrm{m}$).

\vspace{0.2em}
\noindent\textbf{Cosserat rod kinematics.}
The backbone configuration along arc-length $s \in [0,l]$ is described by centreline position $r(s)\in\mathbb{R}^3$ and material frame $R(s)\in SO(3)$. In Cosserat form,
\begin{equation}
\frac{dr(s)}{ds} = R(s)\,v(s),
\qquad
\frac{dR(s)}{ds} = R(s)\,\widehat{u}(s),
\label{eq:cosserat_kinematics}
\end{equation}
where $v(s)\in\mathbb{R}^3$ is the translational strain, $u(s)=[u_x,u_y,u_z]^\top$ is the body-frame curvature--torsion strain, and $\widehat{(\cdot)}$ denotes the skew-symmetric operator. Assuming an inextensible, unshearable section gives the Kirchhoff reduction $v(s)=e_3=[0,0,1]^\top$, hence $dr/ds=R(s)e_3$. In the full Cosserat formulation, $u(s)$ follows from equilibrium and a constitutive law; here it is used only to define the nominal unloaded shape.

\vspace{0.2em}
\noindent\textbf{Boundary conditions and constant-curvature actuation.}
The base boundary conditions at $s=0$ are
\begin{equation}
r(0) =
\begin{bmatrix}
0 & 0 & q_3
\end{bmatrix}^{\top},
\qquad
R(0) = \mathrm{Rot}_z(q_2),
\label{eq:base_bc}
\end{equation}
where $q_3$ sets insertion depth and $q_2$ rotates the bending plane. Assuming constant strain $u(s)=u_0$ along the distal flexible segment:
\begin{equation}
u_0 =
\begin{bmatrix}
\kappa(q_1) & 0 & 0
\end{bmatrix}^{\top},
\qquad
\kappa(q_1) = -\frac{q_1}{2\gamma\,l} \;\; [\mathrm{m}^{-1}],
\label{eq:strain_mapping}
\end{equation}
where $l = 50 \times 10^{-3}~\mathrm{m}$ is the flexible segment length and $\gamma = 1.75 \times 10^{-3}~\mathrm{m}$ is the tendon routing radius. For a symmetric antagonistic tendon pair, this follows from the small-curvature relation $q_1 \approx -2\gamma l \kappa$; axial strain and torsion are neglected.

\vspace{0.2em}
\noindent\textbf{Physics-based simulation and interaction modelling.}
The robot is simulated in PyBullet via a compliant discretisation of the backbone into capsule collision bodies connected by revolute joints. Joint targets encode the nominal constant-curvature shape in Eq.~\ref{eq:strain_mapping}, while joint stiffness $k_j = 2.0 \times 10^{-2}~\mathrm{N\,m\,rad^{-1}}$ and damping $d_j = 5.0 \times 10^{-4}~\mathrm{N\,m\,s\,rad^{-1}}$ provide compliance under external loads. External contact and friction are not imposed as boundary conditions of the reduced-order rod model; instead, they are resolved by the compliant multibody simulation about the nominal shape.

\noindent\textit{Friction and soft contact.}
Bronchoscope--airway contact follows a Coulomb friction model with $\mu_s = 0.3$ and $\mu_d = 0.25$. Normal interaction is modelled by
\begin{equation}
f_n = \max\!\left(0,\,k_n\,\delta + c_n\,\dot{\delta}\right),
\label{eq:soft_contact}
\end{equation}
where $\delta \ge 0$ is penetration depth, $\dot{\delta}$ the relative normal velocity, and $k_n = 4.0 \times 10^{3}~\mathrm{N\,m^{-1}}$, $c_n = 0.8~\mathrm{N\,s\,m^{-1}}$. These values correspond to $\delta = 0.25~\mathrm{mm}$ at $1~\mathrm{N}$ and $\delta = 0.5~\mathrm{mm}$ at $2~\mathrm{N}$, thereby keeping nominal wall penetration sub-millimetre over the relevant load range.

\noindent\textit{Actuation non-idealities.}
To capture transmission delay and control non-idealities, we inject stochastic control delays $\Delta t \sim \mathcal{U}(0, 0.1)\,\mathrm{s}$ and magnitude-dependent scaling terms:
\begin{equation}
q_1(t) = q_{1,\mathrm{cmd}}(t-\Delta t)\left(1 + 0.05\,\frac{|q_{1,\mathrm{cmd}}(t-\Delta t)|}{q_{1,\max}}\right),
\label{eq:noise_q1}
\end{equation}
\begin{equation}
q_2(t) = q_{2,\mathrm{cmd}}(t-\Delta t)\left(1 + 0.05\,\frac{|q_{2,\mathrm{cmd}}(t-\Delta t)|}{2\pi}\right).
\label{eq:noise_q2}
\end{equation}

\subsection{Anatomical Reconstruction and Data Synthesis}

\noindent\textbf{CT Scan Preprocessing.}
Patient-specific anatomical models are extracted from clinical CT scans through an automated segmentation-to-mesh pipeline. We first resample each CT volume to an isotropic grid and apply standard intensity normalisation for lung CT (HU windowing followed by affine normalisation). Airway lumen segmentation is produced using the 3D U-Net architecture from TotalSegmentator~\cite{wasserthal2023totalsegmentator}, which builds on the nnU-Net framework~\cite{isensee2021nnu} with patch-based training and standard geometric and intensity augmentations. The resulting binary mask is converted into a signed distance field (SDF) representation and meshed into a watertight airway surface, which serves both as the rendering geometry and as the input to the medial axis extraction described below.

\noindent\textbf{Automated Data Collection.}
To automatically collect data within the airways, we generate collision-free trajectories by extracting the medial axis from the reconstructed signed distance field (SDF) of the airway geometry (\figref{fig:pipeline}). Starting from surface sample points, we trace along the SDF gradient $\nabla \phi(\mathbf{x})$ in the inward normal direction. Following the grassfire analogy~\cite{tagliasacchi2016skeletons}, trajectories $\mathbf{x}(t)$ propagate inward with $\dot{\mathbf{x}}(t) = -\mathbf{n}(t)$. Medial axis points are identified where the gradient exhibits sign changes, i.e.\ $\frac{d}{dt}[\nabla \phi(\mathbf{x}(t))] \cdot \hat{\mathbf{n}} = 0$, detected via pronounced spikes in the second derivative $\nabla^2 \phi(\mathbf{x})$.
The extracted medial axis forms a navigation graph capturing the airway centreline topology. Trajectory sampling along this skeleton uses adaptive density at bifurcations and high-curvature regions to ensure comprehensive coverage of geometrically complex areas. These medial axis poses serve as collision-free waypoints tracked by an inverse kinematics controller, producing target 6-DoF poses sampled at \SI{10}{\hertz} for rendering photorealistic data streams.

\begin{figure}[t]
    \centering
    \captionsetup{font=footnotesize}
    \begin{footnotesize}
\begin{forest}
  pic dir tree,
  pic root,
  for tree={
    directory,
    fit=band,
    l sep=5mm,
    s sep=0mm,
  },
  [room\_output
    [patient\_001
      [sequence\_001
        [rgb
          [frame\_0000.png, file]
          [$...$, file]
        ]
        [depth
          [frame\_0000.exr, file]
          [$...$, file]
        ]
        [surface_normals
          [frame\_0000.exr, file]
          [$...$, file]
        ]
        [optical_flow
          [frame\_0000.flo, file]
          [$...$, file]
        ]
        [point_clouds
          [frame\_0000.ply, file]
          [$...$, file]
        ]
        [calibration
          [camera\_params.json, file]
        ]
        [metadata
          [trajectory.json, file]
          [timestamps.json, file]
          [robot\_config.json, file]
        ]
      ]
      [sequence\_002
        [$...$, file]
      ]
      [anatomy
        [lung\_model.obj, file]
        [medial\_axis.ply, file]
        [ct\_metadata.json, file]
      ]
    ]
    [patient\_002
      [$...$]
    ]
    [$...$]
  ]
\end{forest}
\end{footnotesize}
    \caption{\textbf{ROOM pipeline output folder structure.} The framework generates synchronized multi-modal sensor data organized by patient anatomy and sequence. Each sequence contains RGB images (600×600), metric depth maps, surface normals, optical flow fields, point clouds, ground-truth poses, and calibration parameters with timestamps.}
    \label{fig:room_output_structure}
\end{figure}
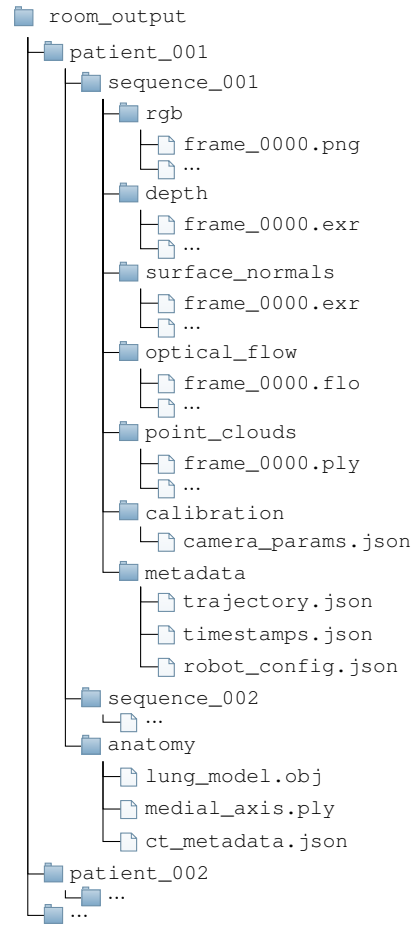

\vspace{0.2em} 
\noindent\textbf{Multi-Modal Data Rendering.} 
For each target pose, we synthesise synchronised data streams: RGB images (600$\times$600), metric and relative depth maps, surface normals, optical flow fields, and point clouds. Each frame is stored together with camera intrinsics/extrinsics, timestamps, and robot configuration to ensure temporal alignment across modalities. Fig.~\ref{fig:room_output_structure} summarises the reconstruction, simulation, and rendered outputs (modalities and metadata). The rendering pipeline utilises Blender’s Principled BSDF shader system with physically-based material properties (base colour, metallic, roughness) to reproduce tissue appearance. We model the directional lighting of the bronchoscope by attaching a point light source with exponential falloff to the tip. For non-RGB modalities we use Blender multi-pass rendering: depth is extracted via the Z-buffer, surface normals are computed from geometry derivatives, and optical flow is computed from inter-frame motion vectors.

\vspace{0.2em} 
\noindent\textbf{Sensor Noise Modelling.} 
Finally, to accurately reproduce noise characteristics of real bronchoscopy RGB images, we employ a frequency-domain system identification approach. Given real endoscopic data $I_{\text{real}}$, we extract the noise component through bilateral filtering as shown in \figref{fig:pipeline}:
\begin{equation}
n_{\text{real}} = I_{\text{real}} - \text{BF}(I_{\text{real}})
\end{equation}
We then analyse the noise spectrum through its Fourier transform $N_{\text{real}}(\omega) = \mathcal{F}\{n_{\text{real}}\}$ and characterize the frequency distribution by the power spectral density $P(\omega) = |N_{\text{real}}(\omega)|^2$. For synthetic data generation, we shape the white noise $w$ to match this spectrum:
\begin{equation}
n_{\text{synth}} = \mathcal{F}^{-1}\left\{\mathcal{F}\{w\} \cdot \sqrt{P(\omega)}\right\}
\end{equation}
The final synthetic RGB image combines the rendered output with the synthesised noise: $I_{\text{synth}} = I_{\text{rendered}} + \beta \cdot n_{\text{synth}}$, where $\beta$ controls the noise amplitude to match medical sensor characteristics. This approach ensures our synthetic data exhibits the same noise statistics as real bronchoscopy imagery, which we observed is crucial for assessing monocular depth estimation performance.

\section{Applications}

We demonstrate ROOM's data for two canonical tasks in medical robotics: multi-view pose estimation and monocular depth estimation evaluation. Additionally, we demonstrate applications of the synthesised data for fine-tuning depth estimation models, as well as potential navigation tasks.

\subsection{Task 1: Multi-View Pose Estimation}

\begin{table}[t]
    \centering
    \captionsetup{font=footnotesize}
    \begin{tabular}{l|ccc}
    \toprule
    \textbf{Method}                         & \textbf{RRA@5°} $\uparrow$ & \textbf{RTA@5°} $\uparrow$ & \textbf{AUC@30°} $\uparrow$ \\
    \midrule
    COLMAP~\cite{colmap2016}       & 41.00 & 0.07 & 16.91 \\
    ORB-SLAM3~\cite{orb_slam3}     & 71.67 & 0.17 & 42.74 \\
    DUSt3R~\cite{duster2024}       & 63.00 & 0.21 & 54.90 \\
    VGGT~\cite{vggt2025}           & 79.00 & 0.25 & 69.09       \\
    \bottomrule
    \end{tabular}
    \caption{\textbf{Comparison of methods across five sequences (Seq0–Seq4)}. Reported values are means across all sequences. Metrics: Relative Rotation Accuracy (RRA@5°), Relative Translation Accuracy (RTA@5°), and Area Under the Curve (AUC@30°). Higher is better (↑).}
    \label{tab:methods_comparison}
\end{table}

The first task is camera pose estimation from multiple views, a fundamental task in medical robotics that underpins downstream bronchoscopy use-cases such as 3D reconstruction. The repetitive branching patterns and limited texture of airways, pose particular challenges for evaluating existing visual odometry and structure-from-motion methods. 

For evaluation, we synthesised realistic reference paths along the airways, to obtain photorealistic RGB images and ground truth poses. 
We evaluated four methods: ORB-SLAM~\cite{orb_slam3} as a classical feature-based baseline, COLMAP~\cite{colmap2016} with sequential matching constraints, and DUSt3R~\cite{duster2024} and VGGT~\cite{vggt2025} as learning-based methods. We measure the Relative Rotation Accuracy (RRA@5°), Relative Translation Accuracy (RTA@5°), and Area Under the Curve (AUC@30°), as done in prior work~\cite{vggt2025}.

Our results are reported in \tabref{tab:methods_comparison}. We report that classical methods achieve only 41\% RRA and 0.07\% RTA tracking success due to insufficient texture, while DUSt3R trained on natural image data reaches 63\% RRA and 0.37\% RTA on held-out sequences. VGGT demonstrates superior performance with 79\% RRA and 0.5\% RTA, representing a substantial improvement over classical approaches. These results align with recent findings in endoscopic domains: ORB-SLAM3 achieves only 25\% frame localisation success on real colonoscopy sequences~\cite{endomapper2023}, while other methods such as CudaSIFT-SLAM shows 70\% improvement over ORB-SLAM3 in colonoscopy mapping coverage~\cite{cudasift2024}. Similarly, pose estimation studies in endoscopy report challenges with classical methods, with specialised endoscopic pose estimation achieving errors of 1.43 mm in bronchoscopy and 3.64 mm in colonoscopy~\cite{structure_depth_endoscopy2024}. The higher performance of VGGT on our bronchoscopy data is consistent with its demonstrated advantages over DUSt3R and traditional methods across multiple benchmarks~\cite{vggt2025}.

\subsection{Task 2: Monocular Depth Estimation}
Monocular depth estimation is an important task in medical robotics~\cite{simcol3d}, since it is challenging to use stereo setups or structured light under the bronchoscope's size constraints---they range from 2.4--6.2 mm in outer diameter~\cite{aats_bronchoscopy}.

We compare different pre-trained depth estimation models using ROOM-generated data. We evaluate four general-purpose foundation models for monocular depth, namely Metric3D-V2~\cite{metric3dv2_2024}, Depth Anything V2 (monocular and relative variants)~\cite{dav2_2024}, and UniDepth (monocular and relative variants)~\cite{unidepth_2024}. Additionally, we evaluate three endoscopy-specialized methods: EndoDAC (transfer from Depth Anything)~\cite{cui2024endodac}, EndoOmni (transfer from DINOv2)~\cite{tian2024endoomni}, and BREA-Depth (transfer from Depth Anything V2)~\cite{brea_depth2025}.
Each model is evaluated using standard depth estimation metrics: L1 error, RMSE, absolute relative error, and delta accuracy thresholds ($\delta < 1.25^i$ for $i \in \{1,2,3\}$).

\begin{table}[t]
    \captionsetup{font=footnotesize}
    \caption{\textbf{Monocular depth estimation results on ROOM synthetic data}. L1 and RMSE are reported in millimetres (mm). $\delta_1$ is the percentage of pixels satisfying $\max(\hat{d}/d, d/\hat{d}) < 1.25$. \dag~Relative-depth methods are scale-aligned per sequence using the ground-truth median depth.}
    \label{tab:depth_results}
    \centering
    \begin{tabular}{lcccc}
    \hline
    \textbf{Method} & \textbf{L1 $\downarrow$} & \textbf{Abs Rel$\downarrow$} & \textbf{RMSE $\downarrow$} & \textbf{$\delta_1$ (\%)$\uparrow$} \\
    \hline
    Metric3DV2~\cite{metric3dv2_2024} & 9.5 & 0.440 & 14.5 & 27.5 \\
    DAV2 (Metric)~\cite{dav2_2024} & 9.7 & 0.459 & 14.7 & 28.5 \\
    DAV2 (Rel.)\dag~\cite{dav2_2024} & 11.3 & 0.486 & 17.9 & 28.2 \\
    EndoDAC~\cite{cui2024endodac} & 9.4 & 0.432 & 14.4 & 29.6 \\
    UniDepth~\cite{unidepth_2024} & 10.6 & 0.476 & 16.6 & 27.1 \\
    EndoOmni~\cite{tian2024endoomni} & 9.2 & 0.428 & 14.2 & 30.2 \\
    BREA-Depth~\cite{brea_depth2025} & 9.1 & 0.421 & 14.1 & 30.8 \\
    \hline
    \end{tabular}
\end{table}

\begin{figure*}[t]
  \centering
  \captionsetup{font=footnotesize}
  \includegraphics[width=1\linewidth]{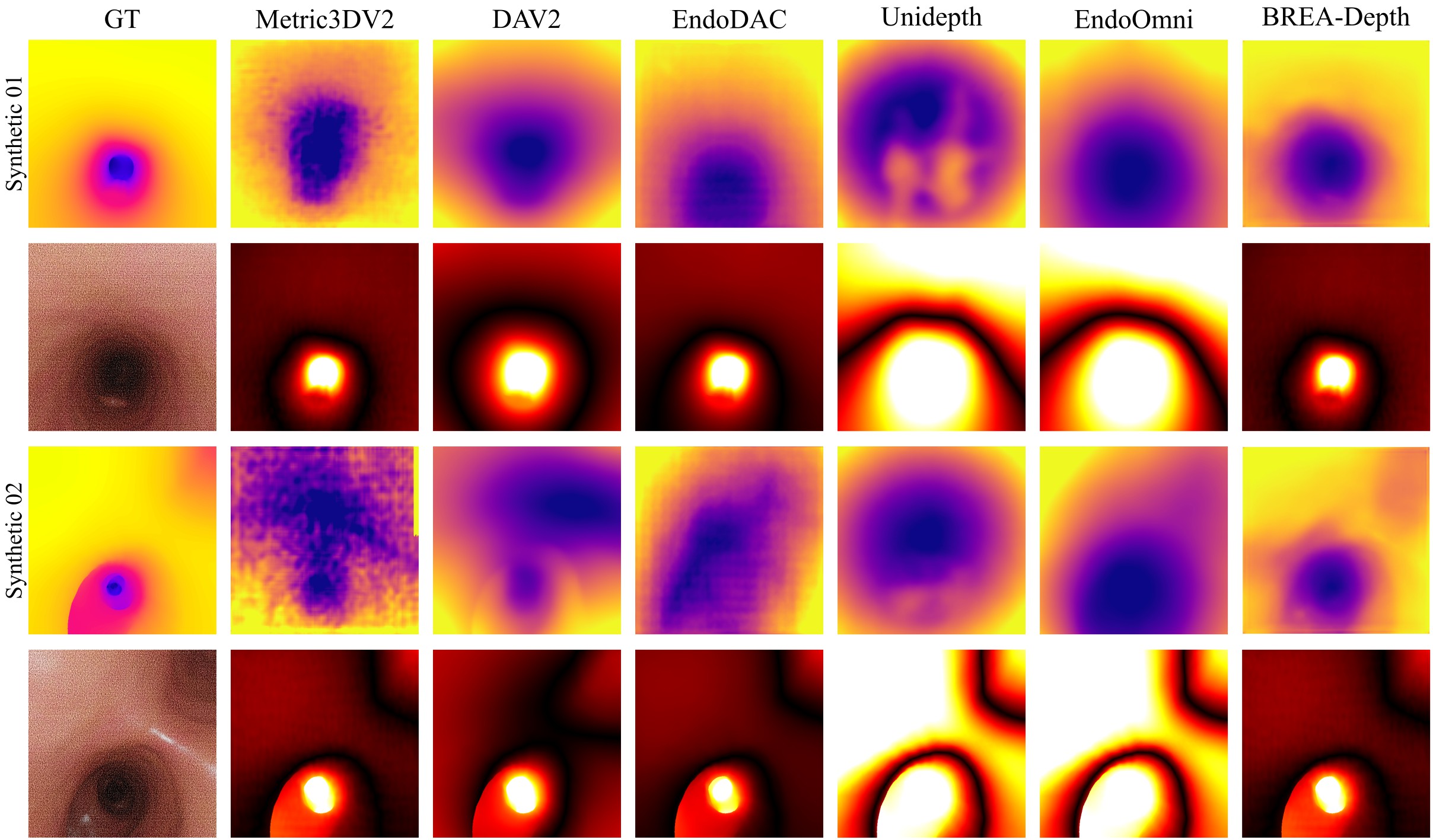}
    \caption{\textbf{Comparative monocular depth estimation results on ROOM synthetic bronchoscopy sequences.} Top rows show L1 error maps between predicted depth estimation and ground truth depth, where warmer colours indicate higher absolute errors, while bottom rows display corresponding RGB inputs with challenging specular highlights and limited texture. Five state-of-the-art models are evaluated: Metric3DV2, Depth Anything V2 (DAV2 Monocular/Relative), Unidepth, EndoOmni, EndoDAC, BREA-Depth, revealing significant performance degradations due to the realistic sensor noise from the simulator and systematic errors concentrated at geometric transitions and specular regions.}
  \label{fig:depth_results}
  \vspace{-10pt}
\end{figure*}

The quantitative results in Tab.~\ref{tab:depth_results} highlight a persistent domain gap for monocular depth in bronchoscopy. Absolute errors remain on the order of centimetres (e.g., L1 $\approx$ 9.0--11.0~mm), while relative performance is weak across all baselines (Abs Rel $\approx$ 0.42--0.49 and $\delta_1 \approx$ 27--31\%), far below the 80--90\% $\delta_1$ commonly reported on natural-image benchmarks. Among all methods, UniDepth achieves the lowest absolute error (L1 $=10.6$~mm, RMSE $=16.6$~mm), while BREA-Depth achieves the best overall relative accuracy (Abs Rel $=0.421$, $\delta_1=30.8\%$), consistent with improved lumen localisation. These results reflect the combined challenges of limited texture, specularities, and extreme depth ranges (2--50~mm) in the bronchoscopy domain.

The error maps in \figref{fig:depth_results} expose systematic failure modes. Errors cluster at specular highlights where wet mucosal surfaces create photometric inconsistencies, and at geometric discontinuities including bifurcations where the tubular structure transitions. Furthermore, DAV2 variants show more diffuse error patterns, while Metric3DV2 and UniDepth maintain better structural coherence but fail at boundaries. The repetitive branching geometry provides insufficient texture gradients for reliable depth cues, particularly evident in the uniform error distribution across smooth airway walls. 

We conclude that the bronchoscopy environment violates core assumptions of existing depth estimation methods, requiring domain-specific training approaches like ROOM's synthetic data generation to bridge this performance gap.

\subsection{Task 3: Fine-tuning Monocular Depth Models}
After the findings of Task 2, we proposed to assess if fine-tuning monocular depth models using synthetic ROOM data could provide performance boosts.
For this task we used three models: the general-use UniDepth and DepthAnything V2 (DAV2), as well as the bronchoscopy-specialised BREA-D. Furthermore, to avoid testing the models using a test set within the same data distribution of the fine-tuning data, we evaluated them on an external bronchoscope dataset with phantom-based depth ground truth~\cite{visentini2017external_data}. We compare their performance before and after fine-tuning using the same depth estimation metrics used in Task 2.

\begin{table}[t]
    \centering
    \captionsetup{font=footnotesize}
    \caption{\textbf{Comparison of original and fine-tuned models on an external bronchoscope dataset.} 
    \textbf{Bold} indicates improvements over the original model after fine-tuning. L1 and RMSE are in millimetres.}
    \scriptsize
    \begin{tabular}{l l cccc}
    \toprule
     & \textbf{Method} & \textbf{L1} $\downarrow$ & \textbf{Abs Rel} $\downarrow$ & \textbf{RMSE} $\downarrow$ & $\delta_1$ (\%) $\uparrow$ \\
    \midrule
    \multirow{3}{*}{\rotlab{Original}}
      & UniDepth~\cite{unidepth_2024} & 8.0 & 0.545 & 10.0 & 19.77 \\
      & DAV2~\cite{dav2_2024} & 20.0 & 0.382 & 24.0 & 42.15 \\
      & BREA-D~\cite{brea_depth2025} & 14.0 & 0.197 & 19.0 & 65.39 \\
    \midrule
    \multirow{3}{*}{\rotlab{Fine-\\tuned}}
      & UniDepth~\cite{unidepth_2024} & \textbf{4.0} & \textbf{0.277} & \textbf{6.0} & \textbf{59.87} \\
      & DAV2~\cite{dav2_2024} & \textbf{15.0} & \textbf{0.291} & \textbf{20.0} & \textbf{55.42} \\
      & BREA-D~\cite{brea_depth2025} & \textbf{13.0} & \textbf{0.192} & \textbf{18.0} & \textbf{67.70} \\
    \bottomrule
    \end{tabular}
    \label{tab:breavsfinetune}
    \vspace{-5pt}
\end{table}

\begin{figure*}[ht]
  \centering
  \captionsetup{font=footnotesize}
  \includegraphics[width=1\linewidth]{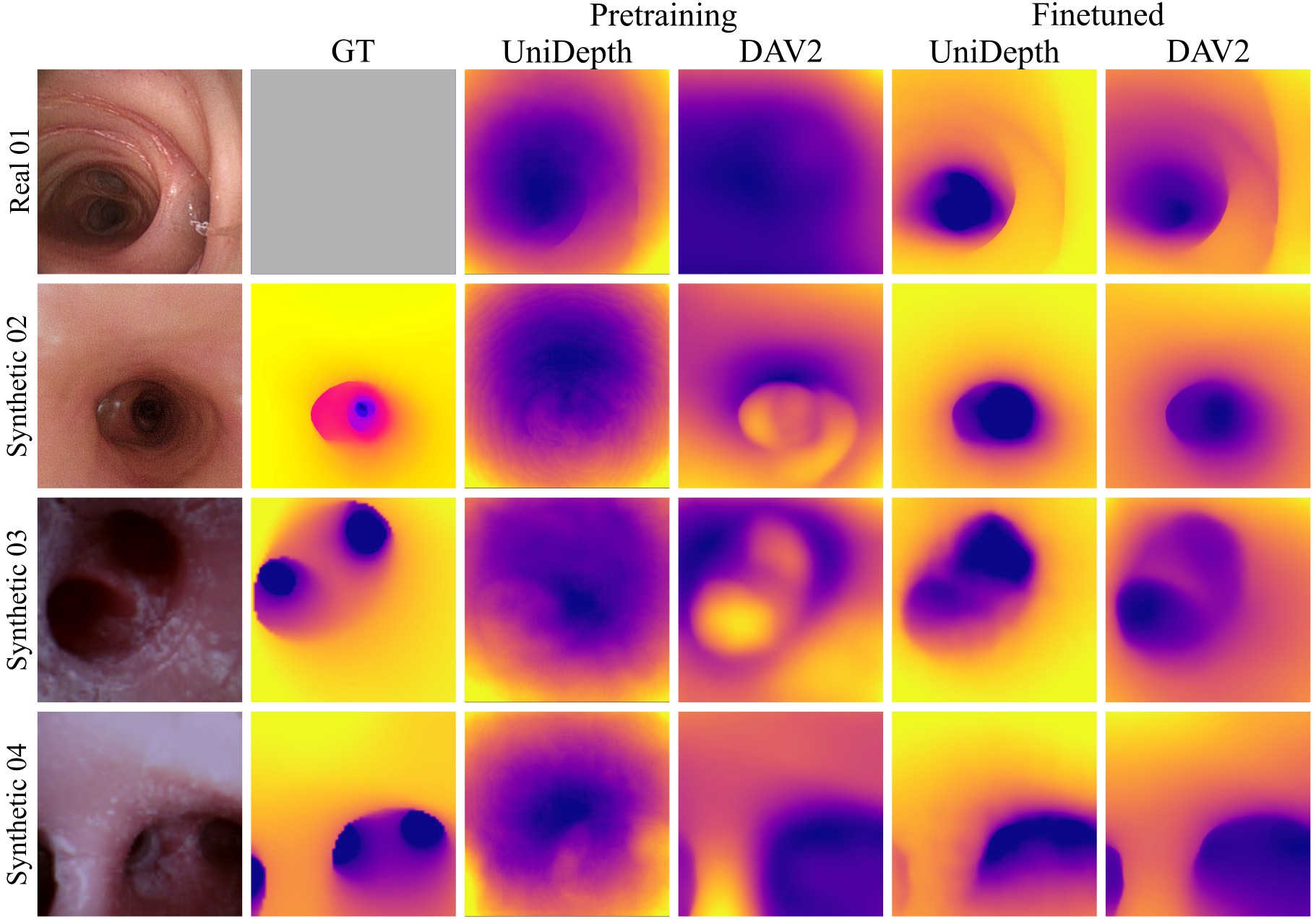}
    \caption{\textbf{Monocular depth estimation examples of pre-trained models and fine-tuned on ROOM.} We show examples on a phantom-based dataset with ground truth~\cite{visentini2017external_data} as well as real images. Please note that the real image does not have depth ground truth available.}
  \label{fig:breavsfinetune}
  \vspace{-10pt}
\end{figure*}

We report the results for the different metrics in \tabref{tab:breavsfinetune} using ten selected representative image-depth pairs. Our results indicate that fine-tuning on synthetic ROOM-generated data improves BREA-Depth across all metrics, with $\delta_1$ accuracy increasing from 65.39\% to 67.70\% (a relative gain of 3.5\%). These improvements are also reflected in qualitative comparisons between the pre-trained and fine-tuned models shown in \figref{fig:breavsfinetune}. We report improvements in the fine-tuned models even when tested on real bronchoscopy images that were part of neither the pre-training nor fine-tuning data.

Our results demonstrate that the synthetic data produced by ROOM provides effective supervision for bridging domain gaps and recovering performance under challenging bronchoscopic conditions, suggesting promising avenues to fine-tune general models in this medical domain.

We evaluate ROOM’s data on two canonical tasks in medical robotics: multi-view pose estimation and monocular depth estimation. We then use the synthetic data for fine-tuning depth estimation models and report transfer to an external bronchoscopy dataset. Finally, we include a qualitative navigation demonstration to illustrate how ROOM’s synchronised modalities (RGB/depth and calibration/poses) can be integrated into a classical planning stack; this demonstration is not presented as a quantitative navigation benchmark.

\begin{figure}[t]
  \centering
  \captionsetup{font=footnotesize}
  \includegraphics[width=1\linewidth, trim=0pt 0pt 0pt 0pt, clip]{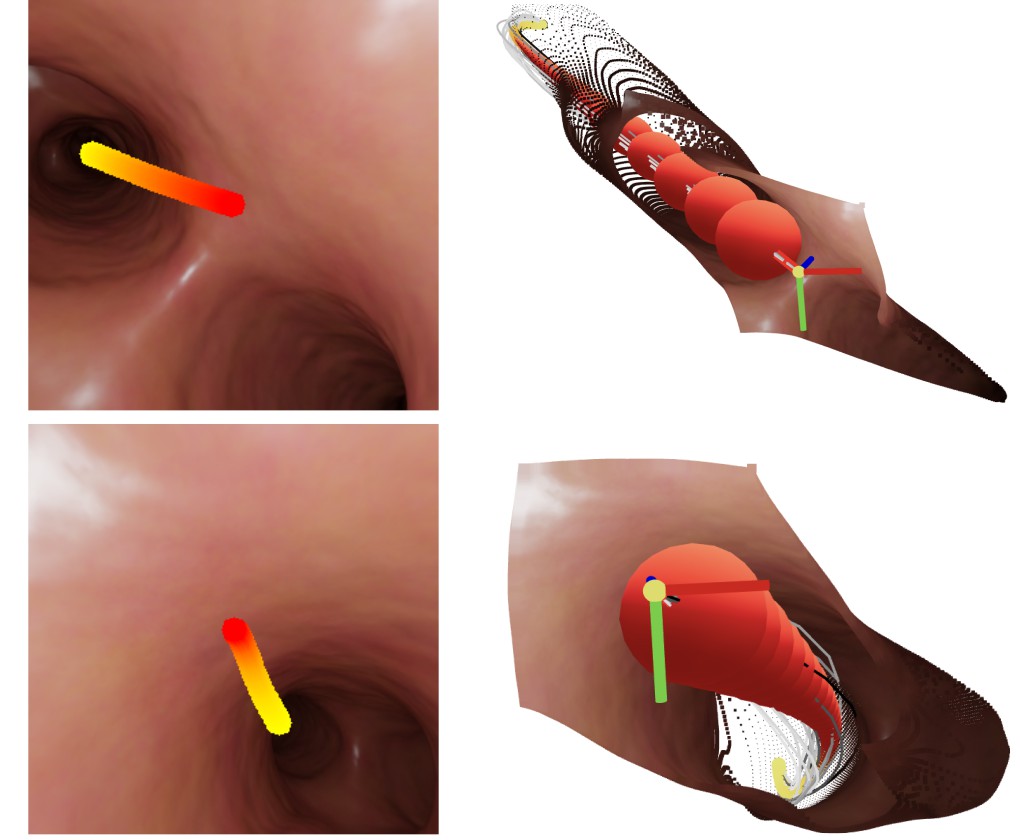}
  \caption{\textbf{Vision-based navigation examples.} We demonstrate qualitative results of the relative monocular depth predictions (scaled with ground-truth scale), as input for a sampling-based local planner. \textit{Left:} projection of the collision-free path. \textit{Right:}  3D visualisation of the path, with the spheres indicating the collision-bodies used by the planner.}
  \label{fig:nav}
\end{figure}

\subsection{Demonstration: Vision-Based Navigation}

Lastly, we provide a qualitative demonstration of how ROOM outputs can be integrated into a vision-based bronchoscope navigation pipeline. The goal is to show that the simulator provides the required synchronised modalities (RGB, depth, and calibration/poses) for downstream planning and control. This is not a quantitative navigation evaluation; therefore do not report navigation success metrics.

We implemented a vision-based navigation method based on a sampling-based planner~\cite{jankowski_vpsto_2023}, using the predicted depth maps to generate a local point cloud map for collision checking. \figref{fig:nav} shows example output paths predicted from single frames, visualising the path from the current camera pose (image centre) to the farthest visible point. The 3D visualisations on the right place the planned path in relation to the robot’s collision geometry (coloured spheres), illustrating feasibility within lumen constraints. This integration shows that models trained or fine-tuned with ROOM data can be used within classical planning stacks. 

\section{Discussion}
Our results show that the synthetic data produced by ROOM can contribute to overcome challenges that well-established methods in multi-view pose estimation and monocular depth estimation face in the bronchoscopy domain. However, there are limitations and aspects for future improvement of the ROOM framework.

First, the anatomical reconstruction pipeline depends on CT scan quality, and may fail with pathological cases exhibiting severe occlusions or abnormal geometries. However, this also presents an opportunity to extend the framework to other endoscopic procedures where CT scans are available, such as colonoscopy and arthroscopy.
Second, while ROOM is built on top of the PyBullet simulator to provide a physically-accurate environment for data collection, it might not fully reflect the contact and deformable dynamics of real bronchia. Enabling support for tissue deformation modelling as well as physiological dynamics such as respiratory motion might also provide realism to the synthesised data.

Lastly, the physical simulator can enable future research in closed-loop navigation systems, enabling its use for validating traditional planners, or for developing imitation learning or reinforcement learning-based navigation policies, as proposed by recent works~\cite{bronchocopilot_2024}. 

\section{Conclusions}

We introduced ROOM, a physics-based simulation framework that addresses the critical data scarcity challenge in bronchoscopy robotics.
By integrating patient-specific anatomical reconstruction, continuum robot physics, and photorealistic rendering at medically relevant scales, ROOM enables generation of diverse training datasets that capture the complexity of clinical procedures. Our evaluation in established tasks such as multi-view pose estimation and monocular depth estimation reveals that the bronchoscopy domain presents significant challenges for existing methods. However, we showed that the synthetic data generated by ROOM can provide avenues for fine-tuning them and improve performance in real settings.

The ROOM framework will be made available for the community. We expect that its modular architecture will enable researchers to test new CT scans, substitute components, or swap rendering engines, physics simulators, or robot models, opening new avenues for medical robotics research.

\balance
\bibliographystyle{IEEEtran}
\bibliography{references}

\end{document}